\pdfoutput=1

\documentclass[11pt]{article}

\usepackage[]{acl}

\usepackage{times}
\usepackage{latexsym}

\usepackage[T1]{fontenc}

\usepackage[utf8]{inputenc}

\usepackage{microtype}

\usepackage{inconsolata}

\usepackage{graphicx}
\usepackage{xspace}
\usepackage{amsmath}
\usepackage{bbm}
\usepackage{multirow}
\usepackage{booktabs}
\usepackage{scalerel}
\usepackage{eqparbox}
\usepackage{algorithm}
\usepackage{algpseudocode}
\usepackage{tabularx}
\usepackage{ragged2e}

\newcommand{\method}[0]{\textsc{Impossible Distillation}\xspace}
\newcommand{\dataset}[0]{DIMPLE\xspace}
\newcommand{\model}[0]{Impossible-T5\xspace}

\newcommand{\eg}{\textit{e.g.,} }
\newcommand{\ie}{\textit{i.e.} }

\newcommand{\teacher}{\mathcal{M}_\mathit{T}}
\newcommand{\student}{\mathcal{M}_\mathit{S}}

\newcolumntype{M}[1]{>{\centering\arraybackslash}m{#1}}
\newcolumntype{Y}{>{\centering\arraybackslash}X}

\title{
Impossible Distillation for Paraphrasing and Summarization: \\
How to Make High-quality Lemonade out of Small, Low-quality Models
}

\author{Jaehun Jung\textsuperscript{$\dagger$} \hspace{.3cm} 
Peter West\textsuperscript{$\dagger$} \hspace{.3cm} 
Liwei Jiang\textsuperscript{$\dagger$} \hspace{.3cm}
Faeze Brahman\textsuperscript{$\dagger\ddagger$} \\
\textbf{Ximing Lu}\textsuperscript{$\dagger$}  \hspace{.3cm} 
\textbf{\hspace{.3cm} Jillian Fisher}\textsuperscript{$\dagger$} \hspace{.3cm}
\textbf{\hspace{.3cm} Taylor Sorensen}\textsuperscript{$\dagger$} \hspace{.3cm}
\textbf{\hspace{.3cm} Yejin Choi}\textsuperscript{$\dagger\ddagger$} \\
\textsuperscript{$\dagger$}Paul G. Allen School of Computer Science \& Engineering, University of Washington\\
\textsuperscript{$\ddagger$}Allen Institute for Artificial Intelligence \hspace{.3cm}  \\
\texttt{hoony123@cs.washington.edu} \\
}

\begin{document}
\maketitle
\begin{abstract}
We present \method, a novel framework for paraphrasing and sentence summarization, that distills a high-quality dataset and model from a low-quality teacher that itself cannot perform these tasks. Unlike prior works that rely on an extreme-scale teacher model (\eg GPT3) or task-specific architecture, we hypothesize and verify the \emph{paraphrastic proximity} intrinsic to pre-trained LMs (\eg GPT2), where paraphrases occupy a proximal subspace in the LM distribution. By identifying and distilling generations from these subspaces, \method produces a high-quality dataset and model even from GPT2-scale LMs. We evaluate our method on multiple benchmarks spanning unconstrained / syntax-controlled paraphrase generation and sentence summarization. Our model with 770M parameters consistently outperforms strong baselines, including models distilled from ChatGPT, and sometimes, even ChatGPT itself. Also, we find that our distilled dataset from 1.5B LMs exhibits higher diversity and fidelity than up to 13 times larger datasets.


\end{abstract}

\section{Introduction}
Training a compact, yet performant model is a non-trivial challenge in modern NLP, even for classical tasks such as paraphrase generation and sentence summarization. While large-scale, high-quality data is central to this goal, human supervision is hard to scale; as such, research efforts have focused on training models with an unsupervised, automatically generated dataset. Common approaches include back-translation \cite{paranmt} and auto-encoding \cite{dae_unsupervised_sentence_compression}, but are often limited in terms of corpus diversity and noisiness \cite{parabank1, parabank2}.

Alternatively, recent works propose to train a compact task model by distilling knowledge from gigantic language models (LLMs) \cite{symkd}. As LLMs such as GPT3 -- often multi-billion scale and instruction-tuned -- are already competent in paraphrasing and summarizing sentences \cite{chatgpt_intent}, a specialized model can be trained by simply imitating LLM generations \cite{chatgpt_intent, inheritsumm}. Despite with limitations (\eg significant budget requirement for data collection), LLM distillation outperforms previous methods without human supervision, giving out an impression that powerful teacher LM is all we need to train a better student.

\begin{figure}[t]
    \centering
    \includegraphics[width=\linewidth]{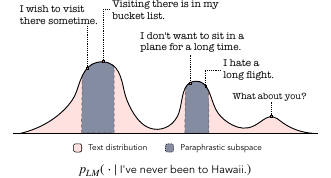}
    \vspace{-15pt}
    \caption{\method develops upon \textit{paraphrastic proximity}: LM's tendency to encode paraphrases on a proximal subspace in its distribution.}
    \label{fig:teaser}
    \vspace{-15pt}
\end{figure}

\begin{figure*}[ht]
    \centering
    \includegraphics[width=.96\textwidth]{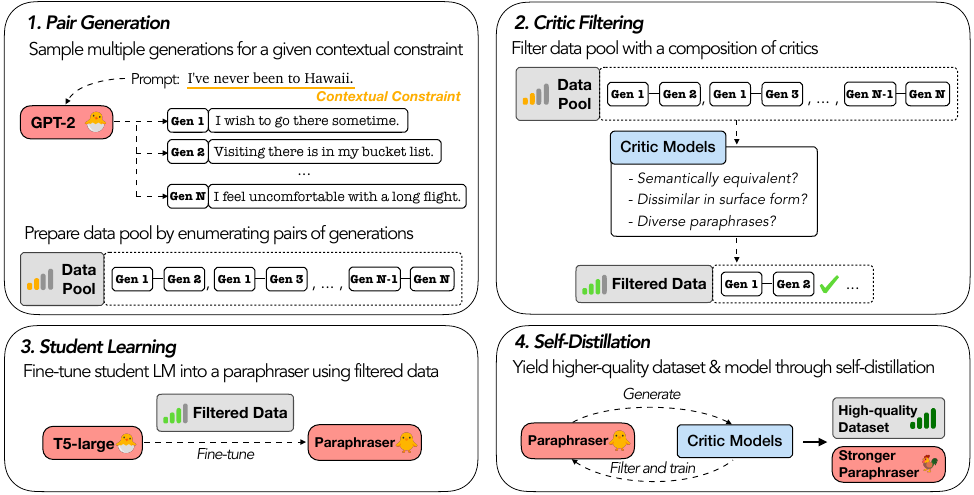}
    \vspace{-5pt}
    \caption{Overview of \method. Starting from low-quality LM (GPT2), we generate a data pool of input-output pairs leveraging paraphrastic proximity, filter it with off-the-shelf critics, and distill a student model on this data pool. By self-distilling the student model, we obtain a high-quality dataset and model for target task.}
    \label{fig:overview}
    \vspace{-10pt}
\end{figure*}

In this work, we envision a seemingly impossible alternative to LLM distillation: instead of an extreme-scale, frontier LLM (\eg GPT3), can we start off with a small, off-the-shelf LM that itself cannot perform paraphrase generation or sentence summarization? We present \method, a novel framework to distill task-specialized dataset and model from GPT2-scale LMs. Our framework requires neither a strong LLM nor human-authored references, yet can distill high-quality paraphrases and summaries comparable to that of prompting the strongest LLMs.

The key observation behind our framework is that a sentence and its paraphrases tend to lie on a proximal subspace in the pretrained LM distribution -- a property we call \textit{paraphrastic proximity} (Fig. \ref{fig:teaser}). In other words, by effectively reducing down the LM search space (\eg by constraining the model with an informative context) toward the paraphrastic subspaces, we can encourage the model to generate multiple sequences that paraphrase each other. As shown in Fig. \ref{fig:overview}, we leverage this property by first constructing a data pool of \textit{(source, paraphrase)} pairs by enumerating a batch of generations sampled given the context. Next, we filter the data pool with off-the-shelf critic models to keep only the pairs with high quality paraphrases, which we subsequently use to fine-tune a student LM. Finally, the student LM is further refined through self-distillation, where the model is trained on its own high-quality paraphrases; as a result, we obtain both a high-quality corpus and a compact, yet powerful model for paraphrasing. Moreover, as \method is grounded on the explicit evaluation of generated pairs, the framework generalizes to sentence summarization by simply re-defining the filters.

Experimental results show that \method is surprisingly effective, both in terms of the distilled data quality and model performance. We first evaluate the quality of our dataset by measuring the semantic fidelity, lexical diversity, and syntactic diversity against three state-of-the-art paraphrase corpora. We find that our dataset, as a purely synthetic corpus generated from 1.5B LMs, shows better metrics in all measures than state-of-the-art datasets: ParaBank \cite{parabank1} that is 13 times larger than ours and ChatGPT-Para \cite{chatgpt_para} generated by orders of magnitude larger ChatGPT. Furthermore, in benchmarks across three distinct tasks -- unconstrained / syntax-controlled paraphrasing and sentence summarization, our model distilled from 1.5B LM outperforms competitive baselines, including both the task-specific methods and the models distilled from ChatGPT \cite{chatgpt}. In human evaluation, our model with 770M parameters is consistently preferred to the ChatGPT-distilled model, and sometimes, even ChatGPT itself.

\section{Paraphrastic Proximity}
\label{sec:paraphrastic_proximity}
We develop \method based on the observation of \textit{paraphrastic proximity} -- \ie when the LM decoding space is constrained with sufficiently informative context, the model can produce multiple generations that paraphrase each other. Notably, \citet{conrpg} indirectly leverages paraphrastic proximity in \textit{context LMs} -- a set of encoder-decoder transformers pre-trained from scratch using specialized training objectives. We, on the other hand, show that paraphrastic proximity holds for off-the-shelf LMs such as GPT2, which we can make use of to distill a high-quality task model and dataset. In this section, we first verify this with GPT2-XL \cite{gpt2}. 


While the exact distribution of all possible generations is intractable, we can obtain an approximation by sampling a large number of generations given a contextual constraint. Concretely, we first sample 1000 context (each with 1-5 sentences) from news articles in XSUM dataset \cite{xsum}. Then we prompt GPT2 to generate 100 next sentences per each contextual constraint. To evaluate whether these sentences are indeed paraphrastic to each other, we measure their (1) pair-wise semantic equivalence, and (2) surface-form dissimilarity. For semantic equivalence, we employ an off-the-shelf NLI model \cite{wanli}, and determine a pair $(x, y)$ to be semantically equivalent if entailment holds in both directions. For surface-form dissimilarity, we compute the Self-BLEU \cite{self_bleu} between sentences, for only the semantically equivalent pairs.

\begin{figure}
    \centering
    \includegraphics[width=.49\textwidth]{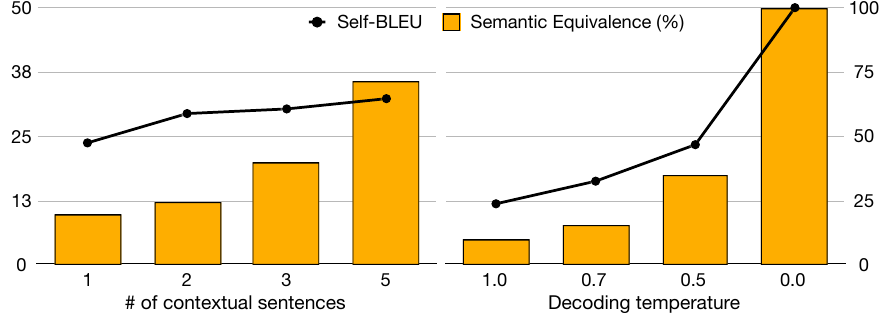}
    \caption{How paraphrastic are GPT2-XL generations? We compute the ratio of semantically equivalent pairs and their average Self-BLEU.}
    \label{fig:pp_chart}
    \vspace{-10pt}
\end{figure}

The results are presented in Fig. \ref{fig:pp_chart}. In the left figure, as the context becomes longer (\ie more informative), the generated sentences are more likely to be semantically equivalent, verifying our assumption. Notably, the high semantic equivalence does not come from merely generating sentences with similar surface form -- even with longer context, the average pair-wise Self-BLEU is around 32\footnote{The average Self-BLEU of human-authored paraphrases in MRPC \cite{mrpc} dataset is 39 \cite{semantic_similarity_prediction}.}. The results indicate that GPT2 can generate a large number of paraphrases simply by over-sampling multiple completions to the given context.

On the right side of Fig \ref{fig:pp_chart}, we also gauge the paraphrastic proximity under various decoding temperatures. Here, the ratio of semantically equivalent pairs dramatically increases as the temperature decreases. Low temperature adjusts the sampling distribution to be more skewed towards regions with high probability mass, hence allowing paraphrases to be more easily sampled from these subspaces. However, it is important that the temperature balance the trade-off between sample efficiency and diversity; when the temperature is too low, the generated sentences are almost identical and hence does not qualify as desirable paraphrase.

\section{Impossible Distillation}
\method starts from an off-the-shelf teacher LM $\teacher$, and distills its knowledge into a student LM $\student$, yielding a specialized model $\mathcal{M}_{\textit{task}}$ for paraphrasing and sentence summarization. As a byproduct of this process, we also obtain a high-quality dataset $\mathcal{D}_{\textit{task}}$. Below, we detail the process focusing on paraphrase generation as the task of interest, then discuss how this generalizes to sentence summarization.


\subsection{Pair Generation}
\label{sec:pair_generation}
We first generate a large pool of candidate \textit{(source-paraphrase)} pairs $\mathcal{C}_{\mathit{T}} = \{(x_1, y_1), \cdots, $ $ (x_{|\mathcal{C}_{\textit{T}}|}, y_{|\mathcal{C}_{\textit{T}}|})\}$ from an off-the-shelf teacher $\teacher$. Our first step is to prepare contextual constraints $c_i$, by sampling 1-5 sentences from $\teacher$:
\begin{equation*}
\begin{split}
    &c_i \sim p_{\teacher}(\cdot)
\end{split}
\end{equation*}
The contextual constraints could be generated either unconditionally or conditioned on a simple prompt (Appendix \ref{sec:implementation_details}). Alternatively, one could sample contextual constraints from human-written corpus (as done in \S\ref{sec:paraphrastic_proximity}). While manually collecting contextual constraints allows fine-grained control over the generation style and domain, we show that LM-generated context suffices to yield a highly diverse and domain-specific data pool without resorting to an external source of data (\S \ref{sec:dataset_evaluation}). 

Next, we generate a batch of next sentences conditioned on each $c_i$, then enumerate candidate pairs as the combinations of these sentences:
\begin{equation*}
\small
\begin{split}
    &\{s_{i1}, \cdots, s_{ik}\} \sim p_{\teacher}(\cdot | c_i; \tau_{\textit{temp}})\\
    &\mathcal{C}_i = \{(s_{im}, s_{in}) | m, n \in [1, k], m \neq n\}
\end{split}
\end{equation*}
Concretely, we set $k = 100$, generating 100 samples per $c_i$ using Nucleus-Sampling \cite{nucleus_sampling}. Based on our preliminary experiments in \S\ref{sec:paraphrastic_proximity}, we set the decoding temperature $\tau_{\textit{temp}} = 0.7$ to balance the diversity and sample efficiency of the generated pairs. Collecting the pairs across all $c_i$s, we obtain the data pool $\mathcal{C}_{\mathit{T}} = \bigcup_i \mathcal{C}_i$.

\subsection{Filtering with Critics}
\label{sec:filtering}
Despite producing a large population of valid paraphrases, our pair generation process is noisy in nature, as it enumerates all possible pairs of generated sentences. For example, Generation 1 (\textit{I wish to visit there sometime.}) and Generation N (\textit{I hate a long flight.}) in Fig. \ref{fig:overview} will constitute a pair in the data pool, although the two sentences have no logical relevance. A crucial step, therefore, is to filter out suboptimal pairs from the data pool and ensure the quality of the distilled dataset.

\paragraph{Semantic Equivalence Filter}
A faithful paraphrase should preserve the semantics of the source statement without hallucinating unsupported content. NLI models are well-suited to quantify this relationship, as they are trained to infer the logical entailment between an arbitrary pair of statements \cite{nli-verifier}. Hence, we define a binary filter using a small NLI model \cite{wanli} as a critic, and discard the pairs that do not achieve the entailment score over the threshold $\tau_{\textit{semantic}}$:
\begin{equation*}
\small
\begin{split}
    f_{\textit{semantic}}(x, y) = \mathbbm{1} \Bigl\{ & p_{\textit{NLI}}(x \Rightarrow y) \ge \tau_{\textit{semantic}} \,\, \land \\
                                                    & p_{\textit{NLI}}(y \Rightarrow x)) \ge \tau_{\textit{semantic}} \Bigr\}
\end{split}
\end{equation*}

\paragraph{Dissimilarity Filter}
A good paraphrase should significantly alter the surface form of the input while preserving its meaning. In \method, the surface-form dissimilarity is achieved by filtering pairs based on (1) the token overlap between sentences and (2) their syntactic difference. For token overlap, we filter the pairs with higher ROUGE-L \cite{rouge} than a threshold $\tau_{\textit{rouge}}$. To gauge the syntactic difference, we follow prior works \cite{sgcp} by first parsing the constituency tree of the source and paraphrase, then filtering based on their tree edit distance (TED):
\begin{equation*}
\small
\begin{split}
    f_{\textit{dissim}}(x, y) = \mathbbm{1} \Bigl\{ & \text{ROUGE}(x, y) \le \tau_{\textit{rouge}} \,\, \land \\
                                                        & \text{TED}(x, y) \ge \tau_{\textit{TED}} \,\, \Bigr\}
\end{split}
\end{equation*}
Intuitively, the two dimensions of dissimilarity complements each other -- while ROUGE filter promotes lexical divergence in each pair, TED filter preempts ``hacking'' the token-overlap metric by simply switching a few words in the source sentence with corresponding synonyms.

\paragraph{Diversity Filter}
Constructing a high-quality corpus is not just about creating valid input-output pairs; ideally, the corpus should cover a diverse range of style and topic within its samples, as the data diversity directly correlates with the robustness of the trained model \cite{augmentation_robustness}. Our data pool might be limited in this regard, as it includes a large number of pairs from the same context $c$, often resulting in multiple pairs having similar $x$ or $y$. To remove the duplicate pairs and promote diversity, we employ an additional critic $f_{\textit{diversity}}$. Concretely, we define two pairs $(x_1, y_1)$ and $(x_2, y_2)$ to be duplicate when one pair entails another, either on the input side ($x_1 \Rightarrow x_2$) or the output side ($y_1 \Rightarrow y_2$). The diversity filter operates by first grouping all entailing pairs, then discarding all but one with the largest entailment score. In practice, this filter can be efficiently implemented using graph traversal; we describe the formal algorithm in Appendix \ref{sec:implementation_details}.

Incorporating all critics, we filter the candidate pool $\mathcal{C}_{\textit{T}}$ into a refined dataset $\mathcal{D}_{\textit{T}}$ as following:
\begin{equation*}
\small
\begin{split}
    \mathcal{D}_{\textit{T}} = \{(x, y)| & (x, y) \in \mathcal{C}_{\textit{T}}, \\
    & f_{\textit{semantic}} \land f_{\textit{dissim}} \land f_{\textit{diversity}}(x,y) = 1\}
\end{split}
\end{equation*}

\subsection{Distilling Student Model}
\label{sec:student_model}
Now that we extracted the paraphrastic knowledge of the teacher $\teacher$ into a dataset $\mathcal{D}_{\textit{T}}$, we use the data to fine-tune the student model into a paraphrase generation model. The student model $\student$ is fine-tuned by maximizing $\mathbbm{E}_{(x,y) \sim \mathcal{D}_{\textit{T}}} [\log p_{\student}(y|x)]$, i.e. the conditional log-likelihood of $y$ given $x$.

Next, the paraphrasing capability of the student is further amplified through self-distillation, by fine-tuning on its own generated high-quality paraphrases. We first sample the input sentence $x$ from the teacher LM $\teacher$, then generate paraphrase $y$ by feeding $x$ into $\student$:
\begin{equation*}
\small
\begin{split}
    \mathcal{C}_{\textit{S}} = \{(x_1, y_1), \cdots | x_i \sim p_{\teacher}(\cdot|c_i); \, y_i \sim p_{\student}(\cdot|x_i)\}
\end{split}
\end{equation*}
Using the same critics as in the previous stage, we filter $\mathcal{C}_{\textit{S}}$ to obtain a high-quality dataset $\mathcal{D}_{\textit{para}}$. Finally, we fine-tune $\student$ on $\mathcal{D}_{\textit{para}}$, yielding the end-stage model $\mathcal{M}_{\textit{para}}$. Consistent with prior findings on self-distillation \cite{revisiting_self_distillation, mysteries}, this simple process significantly improves the performance of our task model, as confirmed by our ablation study (\S\ref{sec:ablation}). In addition, our self-distillation outputs a large-scale, standalone dataset $\mathcal{D}_{\textit{para}}$ that can be evaluated and reused, \eg to directly train a paraphrasing model without re-iterating the distillation procedure.

\begin{table*}[t]\centering
\resizebox{.96\textwidth}{!}{
    \begin{tabular}{lccccccc}\toprule
         \multirow{2.5}{*}{\textbf{Dataset (\# Instances)}} & \textbf{Semantic Similarity} & \multicolumn{4}{c}{\textbf{Lexical Diversity}}& \multicolumn{2}{c}{\textbf{Syntactic Diversity}} \\\cmidrule(lr){2-2}\cmidrule(lr){3-6}\cmidrule(lr){7-8}
         & Cosine Sim. $\uparrow$ & $H_2$ $\uparrow$ & $H_3$ $\uparrow$ & MSTTR $\uparrow$ & Jaccard Sim. $\downarrow$ & TED-3 $\uparrow$ & TED-F $\uparrow$ \\\midrule
         ParaBank1 (57.0M) & 81.77 & 17.07 & 21.66	& 45.52 & 48.41 & 3.59 & 14.53 \\
         ParaBank2 (19.7M) & 82.50	& 17.48 & \underline{21.44} & \underline{46.16} & \textbf{43.44} & 4.04 & 17.41 \\
         ChatGPT-Para (2.1M) & 85.44 & \underline{17.67} & 21.41 & 35.83 & 44.56 & \underline{4.26} & \underline{20.15} \\
         \dataset (4.2M) & \textbf{87.68} & \textbf{17.75} & \textbf{22.46} & \textbf{53.08} & \underline{43.62} & \textbf{5.02} & \textbf{29.84} \\\bottomrule
    \end{tabular}
}
\caption{Quality comparison between paraphrase datasets. \dataset, as a purely synthetic corpus generated from 1.5B LMs, exhibits better diversity compared to others, including the dataset constructed by prompting ChatGPT.}
\label{tab:dataset_evaluation}
\end{table*}


\subsection{Endowing Controllability}
\label{sec:controllability}
Recent works emphasize the importance of syntactic control in paraphrase generation, allowing the model to generate an output paraphrase tailored to users' need \cite{paranmt-small}. In \method, endowing controllability to the student model is straightforward. We first prepare the dataset $\mathcal{D}_{\textit{control}}$ by parsing the constituency tree $t$ of each paraphrase $y$ in $\mathcal{D}_{\textit{para}}$: 
\begin{equation*}
\small
\begin{split}
    \mathcal{D}_{\textit{control}} = \{(x,y,t)|(x,y) \in \mathcal{D}_{\textit{para}}, t = \text{parse}(y)\}
\end{split}
\end{equation*}
Then a controllable model $\mathcal{M}_{\textit{control}}$ can be trained, by fine-tuning $\student$ to generate $y$ given the source $x$ and the tree $t$.

\subsection{\dataset and \model}
\label{sec:dataset_and_model}
To test \method in both general and domain-specific paraphrasing, we use two teacher LMs -- GPT2-XL and BioGPT (\citealp{biogpt}; for biomedical domain), all with 1.5B parameters. We use T5-large \cite{t5} with 770M parameters as our student LM, and train it with 400k filtered pairs distilled from the teacher models. After self-distilling the student model, we yield a specialized paraphrase generation model we call \textbf{\model}, along with a large-scale corpus with 4M high-quality pairs (2M for general domain and 2M for biomedical domain). We name this dataset \textbf{\dataset}\footnote{\textbf{D}ataset of \textbf{Imp}ossib\textbf{le} Paraphrases.}. Additional implementation details such as generation parameters and filter thresholds are provided in Appendix \ref{sec:implementation_details}.

\subsection{Sentence Summarization}
\label{sec:sentence_summarization}
In addition, we generalize \method for abstractive sentence summarization, a task akin to paraphrasing but with different goals \cite{context_matching}. Whereas paraphrase generation searches for an alternative form of the source sentence while preserving all its information, summarization aims for a succinct representation of the given sentence, at the expense of losing tangential details. In our method, the distillation process is grounded on the explicit evaluation of generated pairs, hence the framework can generalize to summarization by simply redefining the filters. For example, one could add a length filter to the critics, guaranteeing that the output is strictly shorter than the input in the filtered pairs. In Appendix \ref{app:generalization_to_sentence_summarization}, we describe the details of the generalized pipeline for sentence summarization.

\section{Experiments}
\begin{table*}[t]\centering
\resizebox{.99\textwidth}{!}{
    \begin{tabular}{lcccccccccccc}\toprule
         \multicolumn{1}{c}{\textbf{Dataset}} & \multicolumn{4}{c}{\textbf{MRPC}}& \multicolumn{4}{c}{\textbf{ParaNMT-small}} & \multicolumn{4}{c}{\textbf{ParaSCI-arXiv}}\\\cmidrule(lr){2-5}\cmidrule(lr){6-9}\cmidrule(lr){10-13}
         \multicolumn{1}{c}{\textbf{Model}} & \textit{iBLEU} & \textit{B-iB} & \textit{BLEU} & \textit{R-L} & \textit{iBLEU} & \textit{B-iB} & \textit{BLEU} & \textit{R-L} & \textit{iBLEU} & \textit{B-iB} & \textit{BLEU} & \textit{R-L} \\\midrule
         Copy-Input & 1.1 & 0.0 & \textbf{44.4} & \textbf{65.7} & -13.7 & 0.0 & \textbf{23.3} & \textbf{38.2} & 0.0 & 0.0 & \textbf{42.8} & \textbf{60.2} \\\midrule
         GPT-3.5 & 5.7 & 66.5 & 18.1 & 35.7 & 6.2 & 79.6 & 14.1 & 32.2 & \textbf{4.0} & 63.2 & 16.2 & 37.5 \\
         ChatGPT & \textbf{5.9} & \textbf{67.7} & 18.0 & 36.5 & \textbf{6.7} & \textbf{81.3} & 13.4 & 32.8 & 3.6 & \textbf{63.9} & 16.4 & 36.1 \\\midrule
         $\text{T5}_{\text{ParaBank1}}$ & 5.8 & 60.0 & 27.3 & 50.3 & 4.4 & 70.3 & 19.2 & 36.5 & 3.1 & 53.3 & 26.2 & 48.6 \\
         $\text{T5}_{\text{ParaBank2}}$ & 6.1 & 60.5 & 27.7 & 51.4 & 5.3 & 71.5 & 19.7 & 37.6 & 3.7 & 55.1 & 24.9 & 48.2 \\
         $\text{T5}_{\text{ChatGPT-Para}}$ & 5.4 & 62.9 & 21.2 & 40.7 & 5.2 & 74.4 & 15.2 & 33.5 & 3.8 & 59.6 & 23.8 & 41.0 \\
         \model & \textbf{7.3} & \textbf{67.1} & 26.0 & 46.8 & \textbf{5.9} & \textbf{79.7} & 16.3 & 31.8 & \textbf{4.3} & \textbf{64.5} & 25.6 & 44.9 \\\bottomrule
    \end{tabular}
}
\caption{Experimental results of \model and baselines on unconstrained paraphrase generation. \model outperforms the same size model trained on much larger datasets, and is competitive to 175B LLM in both general and domain-specific benchmarks.}
\label{tab:unconstrained_performance}
\end{table*}


\subsection{Dataset Evaluation}
\label{sec:dataset_evaluation}
\paragraph{Evaluation Setup} First, we directly compare the quality of \dataset against three large-scale paraphrase corpora: ParaBank1 \cite{parabank1}, ParaBank2 \cite{parabank2}, and ChatGPT-Para \cite{chatgpt_para}. Both ParaBank1 and ParaBank2 are based on back-translation; ParaBank1 imposes lexical constraints to promote the diversity of paraphrases, and ParaBank2 additionally clusters and resamples generations to further improve the syntactic diversity. ChatGPT-Para is a dataset distilled from ChatGPT, by instructing the LLM to paraphrase sentences from Quora \cite{quora}, SQUAD 2.0 \cite{squad2.0} and CNN/DM \cite{cnn_dm}.

Following \citet{paraamr}, we measure the semantic similarity, lexical diversity and syntactic diversity of each dataset. For semantic similarity, we compute the average cosine-similarity between source and paraphrase measured by SimCSE \cite{simcse}. To estimate lexical diversity, we use 2/3-gram entropy, mean-segmented token type ratio (MSTTR; \citealp{lexical_richness}), and the token-level Jaccard similarity between source and paraphrase. For syntactic diversity, we compute the average pairwise tree edit distance, either for the top 3 layers (TED-3) or the full tree (TED-F).

\paragraph{Results} The results are shown in Table \ref{tab:dataset_evaluation}. In all 3 dimensions, \dataset consistently outperforms all baseline datasets. Notably, this includes ParaBank1 that is more than 13 times larger than \dataset, demonstrating the sample efficiency of our framework in extracting diverse paraphrastic knowledge. In addition, the superior results of our dataset compared to ChatGPT-Para implies that the scale of the LM is not the only factor that determines the quality of generated data. By effectively constraining the LM search space and filtering pairs with a composition of critics, \method makes it possible to generate a high-quality dataset from a small, low-quality LM.

\begin{table}[t]
\centering
\small
\begin{tabular}{lcccc}\toprule
    \textbf{Model} & \textbf{Fleunt} & \textbf{Faithful} & \textbf{Dissimilar} \\\midrule
    ChatGPT & \textbf{2.8} & 2.37 & \underline{2.38} \\
    $\text{T5}_{\text{ParaBank1}}$ & 2.45 & 2.33 & 2.09 \\
    $\text{T5}_{\text{ParaBank2}}$ & 2.47 & \underline{2.50} & 2.21 \\
    $\text{T5}_{\text{ChatGPT-Para}}$ & 2.64 & 2.30 & 2.33 \\\midrule
    \model & \underline{2.74} & \textbf{2.55} & \textbf{2.40} \\\bottomrule
\end{tabular}
\caption{Human evaluation results (\citeauthor{krippendorff}'s alpha = 0.62; substantial inter-annotator agreement). To minimize subjectivity, we use strict 3-level scale, where 3 indicates perfect satisfaction, and 1 indicates complete dissatisfaction of the desired property. }
\label{tab:unconstrained_human_eval}
\vspace{-5pt}
\end{table}

\subsection{Unconstrained Paraphrase Generation}
\paragraph{Evaluation Setup} In this section, we evaluate \model in multiple benchmarks for paraphrase generation without syntactic control. We use three human-curated benchmarks spanning general and domain-specific paraphrase generation: MRPC \cite{mrpc}, ParaNMT-small \cite{sgcp}, and ParaSCI-arXiv \cite{parasci}. For baselines, we fine-tune T5-large with ParaBank1, ParaBank2 and ChatGPT-Para, the three large-scale paraphrase corpora analyzed in \S\ref{sec:dataset_evaluation}. We also consider LLM-based baselines, by zero-shot prompting GPT3.5 (\textit{text-davinci-003}; \citealp{gpt3.5}) and ChatGPT.

\paragraph{Results} In Table \ref{tab:unconstrained_performance}, we report BLEU and ROUGE-L (R-L) along with iBLEU \cite{iBLEU} and BERT-iBLEU (B-iB), a metric known to better correlate with human judgements of paraphrase quality \cite{dynamic_blocking}. Consistent to the prior findings on the brittleness of token-overlap metrics \cite{bertscore}, BLEU and ROUGE-L fail to accurately assess the paraphrase quality. In fact, a simple baseline that merely copies the input as an output (Copy-Input) marks state-of-the-art on these metrics, across all datasets. 

A clearer tendency is shown with iBLEU and BERT-iBLEU: \model consistently outperforms the same size model trained on order of magnitude larger ParaBank, showing up to 10\% relative improvement across all benchmarks. Moreover, \model is the only 770M model comparable to 175B GPT-3.5 across all benchmarks. Notably in an expert domain (ParaSCI), it even outperforms ChatGPT. Additional results against state-of-the-art unsupervised paraphrase generation methods are presented in Appendix \ref{sec:additional_experiments}.

\paragraph{Human Evaluation} We additionally conduct human evaluation to compare the quality of the generated paraphrases. We generate 200 paraphrases with each model using MRPC corpus, and ask six Mechanical Turk workers to evaluate whether each paraphrase is (1) fluent, (2) faithful to the source, and (3) dissimilar to the source (Appendix \ref{sec:human_evaluation_details}). Table \ref{tab:unconstrained_human_eval} shows the results. Consistent to the quantitative metrics, human annotators prefer paraphrases from \model than the competitive baselines. We find that our model is generally considered to be more faithful to the original statement than ChatGPT while sufficiently altering the surface form. Notably, the high faithfulness and dissimilarity does not come from sacrificing the soundness of generation, marking better fluency score than both $\text{T5}_{\text{ParaBank}}$ and $\text{T5}_{\text{ChatGPT-Para}}$.

\subsection{Syntactically Controlled Paraphrase Generation}
\begin{table}[t]
\centering
\resizebox{.48\textwidth}{!}{
    \begin{tabular}{lcccc}\toprule
        \textbf{Model} & \textbf{iBLEU} $\uparrow$ & \textbf{B-iB} $\uparrow$ & \textbf{R-L} $\uparrow$ & \textbf{TED-F} $\downarrow$  \\\midrule
        $\text{ChatGPT}_{\text{0-shot}}$ & 9.1 & 85.8 & 41.6 & 11.6 \\
        $\text{ChatGPT}_{\text{5-shot}}$ & 9.0 & \underline{85.9} & 42.2 & 10.3 \\
        $\text{T5}_{\text{ParaBank1}}$ & 10.7 & 82.3 & \underline{55.6} & \textbf{8.4} \\
        $\text{T5}_{\text{ParaBank2}}$ & \underline{10.9} & 84.7 & \textbf{57.5} & 8.8 \\
        $\text{T5}_{\text{ChatGPT-Para}}$ & 10.5 & 79.4 & 47.6 & 10.4 \\\midrule
        \model & \textbf{11.2} & \textbf{86.6} & 51.8 & \underline{8.5} \\\bottomrule
    \end{tabular}
}
\caption{Results on syntactically controlled paraphrase generation. \model outperforms baselines in both paraphrase quality and controllability.}
\label{tab:constrained_performance}
\end{table}

\paragraph{Evaluation Setup} Next, we assess \model in syntactically controlled paraphrase generation. We use ParaNMT-small, where each sample consists of a source $x$, a syntactic exemplar $z$, and a paraphrase $y$ of $x$ that follows the syntax of $z$. Since the controllable version of our model is trained with the constituency tree as input, we first parse $z$ and feed the tree into our model (along with $x$) during inference. For baselines, we consider T5 trained with existing corpora, additionally annotated with the tree of target paraphrases. We also prompt ChatGPT to generate paraphrase using the same syntax with $z$.

\paragraph{Results} The results are shown in Table \ref{tab:constrained_performance}. \model outperforms baselines across all metrics except ROUGE-L. Notably, the syntax conformity of ChatGPT is substantially poor, even with 5-shot examples of syntax-controlled paraphrases. The results imply that distilling a fine-grained controllable model could be a reasonable alternative to prompting LLM with a textual description of the desired output.

\begin{table}[t]
\centering
\small
\resizebox{.48\textwidth}{!}{
    \begin{tabular}{lccccc}\toprule
        \multirow{2.5}{*}{\textbf{Model}} & \multicolumn{2}{c}{\textbf{Automatic}} & \multicolumn{3}{c}{\textbf{Human}} \\\cmidrule(lr){2-3}\cmidrule(lr){4-6}
        & B-F1 & R-L & Fluent & Faithful & Concise  \\\midrule
        ChatGPT & \underline{84.8} & \textbf{33.6} & \textbf{2.55} & \underline{2.44} & 2.32 \\
        Referee & 78.2 & 29.2 & 2.45 & 2.33 & \underline{2.41} \\
        $\text{T5}_{\text{ParaBank}}$ & 77.5 & 29.6 & 2.21 & 2.17 & 1.96 \\\midrule
        \model & \textbf{85.1} & \underline{30.3} & \underline{2.46} & \textbf{2.53} & \textbf{2.49} \\\bottomrule
    \end{tabular}
}
\caption{Results on sentence summarization. We report BERTScore-F1 \cite{bertscore} and ROUGE-L for automatic evaluation. In addition, six crowd-source workers qualitatively assessed the 100 summaries per each model with 3-level likert scale. }
\vspace{-10pt}
\label{tab:summarization}
\end{table}

\subsection{Sentence Summarization}
\paragraph{Evaluation Setup} We use Gigaword \cite{gigaword}, a representative benchmark for sentence summarization. For baselines, we use ChatGPT and Referee \cite{referee}, an unsupervised summarizer distilled from GPT3. We also train T5-large on $\text{ParaBank}_{\text{summ}}$, a variant of ParaBank2 filtered using the same set of summarization critics as for \method.

\paragraph{Results} The results are as seen in Table \ref{tab:summarization}. We observe that re-purposing a paraphrase corpus for summarization ($\text{T5}_{\text{ParaBank}}$) leads to sub-optimal performance, as the back-translation does not reflect the concise nature of summaries. In contrast, the critic models in \method explicitly participate in the data-generating process, by promoting the model to generate outputs that satisfy the desired properties of critic models. As a result, \method successfully generalizes to summarization, only by plugging in the redefined composition of filters.

\begin{table}[t]
    \centering
    \small
    \begin{tabular}{lc}\toprule
        \textbf{Model} & \textbf{BERT-iBLEU} \\\midrule
        Student model w/o Self-distillation & 64.0 \\
        $\text{T5}_{\text{ChatGPT-Para}}$ & 62.9 \\
        $\text{T5}_{\text{ChatGPT-Para}}$ + Self-distillation & 63.3  \\
        $\text{T5}_{\text{ChatGPT-Para}}$ + Critic Filtering & 64.1 \\ \midrule
        \model & \textbf{67.1} \\\bottomrule
    \end{tabular}
    \caption{Ablation study on MRPC. The best configuration is \method incorporating both critic filtering and self-distillation.}
    \vspace{-10pt}
    \label{tab:ablation}
\end{table} 

\subsection{Ablation Study}
\label{sec:ablation}
In Table \ref{tab:ablation}, we conduct an ablation study to analyze the contribution of different components in \method.

\begin{table*}[htb!]\centering
    \resizebox{\textwidth}{!}{\small
    \begin{tabularx}{\textwidth}{M{0.12\textwidth}m{0.83\textwidth}}\toprule
        \multicolumn{2}{c}{\textbf{Paraphrase Generation (General domain)}}\\
        \midrule\midrule
        \Centering{Constraint $c$} & As part of the process for the upcoming release of the Android M, Google is also adding a new camera API to the latest Android OS.\\
        \midrule
        \Centering{Sentence $x$} & This API allows third-party apps to use the camera of Android devices.\\
        \midrule
        \Centering{Paraphrase $y$} & The new API will allow developers to use Android's camera features to create custom apps.\\
        \bottomrule\vspace{3pt}
    \end{tabularx}
    }
    \resizebox{\textwidth}{!}{\small
    \begin{tabularx}{\textwidth}{M{0.12\textwidth}m{0.83\textwidth}}\toprule
        \multicolumn{2}{c}{\textbf{Paraphrase Generation (Biomedical domain)}}\\
        \midrule\midrule
        \Centering{Constraint $c$} & The impact of obesity on health-related quality of life (HRQOL) in adolescents and young adults with spinal deformity is not well described.\\
        \midrule
        \Centering{Sentence $x$} & The purpose of this study was to compare HRQOL measures in adolescent idiopathic scoliosis (AIS) patients with and without obesity.\\
        \midrule
        \Centering{Paraphrase $y$} & This study aimed to investigate the relationship between HRQOL and obesity in adolescents with idiopathic scoliosis (AIS).\\
        \bottomrule\vspace{3pt}
    \end{tabularx}
    }
    \resizebox{\textwidth}{!}{\small
    \begin{tabularx}{\textwidth}{M{0.12\textwidth}m{0.83\textwidth}}\toprule
        \multicolumn{2}{c}{\textbf{Summarization (General domain)}}\\
        \midrule\midrule
        \Centering{Constraint $c$} & There had been fears the flare could ignite the escaping gas at the Elgin platform, about 150 miles (240 km) east of the Scottish city of Aberdeen, potentially causing a huge explosion. Total said it had received the first indication that the flare might be out at lunchtime on Friday. The firm is ``mobilizing all means to allow these options to be implemented," it said. The company, which is still investigating the cause of the leak, estimates that 200,000 cubic meters of gas a day are escaping.\\
        \midrule
        \Centering{Sentence $x$} & ``The gas cloud is fairly small in size and prevailing winds are blowing it away from the platform and dispersing it,'' Total said.\\
        \midrule
        \Centering{Summary $y$} & The gas cloud is small and blowing away, Total said.\\
        \bottomrule\vspace{3pt}
    \end{tabularx}
    }
    \resizebox{\textwidth}{!}{\small
    \begin{tabularx}{\textwidth}{M{0.12\textwidth}m{0.83\textwidth}}\toprule
        \multicolumn{2}{c}{\textbf{Summarization (Biomedical domain)}}\\
        \midrule\midrule
        \Centering{Constraint $c$} & A banana primarily consists of carbo hydrate chains (sugar), but also contains some minor amount of minerals and vitamins. Let's see what happens with this stuff - Sugar: Will be broken down to either be stored as fat (another form of carbo hydrate chains) or broken up and used to provide cell energy; the resulting "waste" hydrogen and carbon is disposed of in form of CO2 or H2O. Minerals: Are mainly used to regenerate organs/tissue and other organ functions; these could probably be still in your body, but even if they are, they are probably very rare. Vitamins: The atoms are very often disposed after use, so they too leave your body. \\
        \midrule
        \Centering{Sentence $x$} & They do leave in rather short time frames, because the body can't store them well and needs it daily (that is why your diet should include them).\\
        \midrule
        \Centering{Summary $y$} & They do leave in a short time, as the body does not store them long.\\
        \bottomrule
    \end{tabularx}
    }
    \caption{Qualitative examples of pair generation. Along with each $x$ and $y$, we present contextual constraint $c$ used for pair generation.}
    \label{tab:qualitative_examples}
\end{table*}

\paragraph{Does self-distillation matter?}
We analyze the contribution of self-distillation in two ways. First, we omit the self-distillation stage in our framework and directly test the student model $\student$ distilled from $\teacher$. BERT-iBLEU in this case degrades by 3.1 from \model, indicating the importance of self-distillation in amplifying model capability. Next, in order to verify whether self-distillation is a dominant contributor to the performance, we iterate self-distillation on $\text{T5}_{\text{ChatGPT-Para}}$. While the performance of $\text{T5}_{\text{ChatGPT-Para}}$ gets better with iterative distillation, the improvement is relatively small, leading to worse performance than our model without self-distillation. The result confirms that while self-distillation helps in improving the end-stage performance, the diversity of data distilled from teacher model is crucial to fully elicit the student model's capability.

\paragraph{Is it all about critics?}
At the core of \method are the critic models, filtering the noisy data pool generated from the teacher and aligning it for the target task. Therefore, it would be reasonable to ask whether the performance of \model solely comes from the composition of critics in our framework. To methodically verify this, we filter ChatGPT-Para using the same set of critics as in our framework, and train T5-large on the filtered dataset. 

In this configuration, BERT-iBLEU on MRPC marks 64.1, improving over the original ChatGPT-Para but still falling behind \model. We attribute this to the relatively small size of the filtered dataset $(n \approx 340k)$, primarily due to a large portion of pairs not passing either the Dissimilarity or Diversity Filter. While ChatGPT can generate sensible paraphrases, it is not aligned with the specific evaluation criteria defined in the filtering stage, leading to the poor sample efficiency. Although the issue maybe mitigated via fine-tuning the teacher or over-sampling more generations, such solution would require substantially more compute than GPT2-scale LMs. In this sense, our framework provides an attractive alternative to LLM distillation, incorporating a small, cost-efficient data generator and a composition of filters, in replace of a gigantic data generator.

\section{Related Works}
\paragraph{Unsupervised Paraphrasing and Summarization}
Conventional approaches for unsupervised paraphrasing and summarization have focused on task-specific surrogates -- \eg back-translation \cite{paranmt, parabank2} and autoencoding \cite{synpg, seq3} -- that guide the model toward desired output. These surrogate tasks inherently provide weak supervision signal compared to the complexity of the target task, often mandating carefully engineered perturbations \cite{paraamr, dynamic_blocking}, auxiliary constraints \cite{paramac, mcpg} or a complete re-training of teacher model \cite{conrpg}. Apart from the task-specific methods, a growing line of research seeks to harness LLMs to paraphrase and summarize without supervision \cite{llm_paraphrasing, gpt3_summarization}. In fact, recent findings suggest that zero-shot generations prompted from LLMs exhibit human-level quality in various use-cases \cite{machine_paraphrase_plagiarism}.

\paragraph{Task-solving with Language Model}
More broadly, task-solving capabilities of LMs have been tested and analyzed across domains \cite{mmlu}. While large-scale pre-training allows models to acquire sufficient knowledge to solve complex tasks \cite{knowledge-worker, plasma, m-prompting}, recent works suggest that their full capability is elicited from aligning the model knowledge with additional fine-tuning -- e.g. using instruction data \cite{flan-t5, super-naturalinstructions} and human feedback \cite{gpt3.5, finetuning_human_preference} -- which often requires a curated set of annotated data. Our work suggests an alternative to this paradigm, by identifying and leveraging the paraphrastic knowledge intrinsic to the LM, rather than human annotation.

\paragraph{Data Generation with Language Model}
Another line of related works propose to directly distill models with LM-generated data, improving model reasoning \cite{star, distilling-step-by-step}, robustness \cite{rationalization_robustness}, controllability \cite{referee}, and language understanding \cite{zerogen}. These works essentially follow the conceptual framework of Symbolic Knowledge Distillation \cite{symkd}, where a teacher model's knowledge is transferred to a student model via a symbolic, textual dataset. Other works explore to extract a standalone corpus from LMs, whether it be knowledge base \cite{prompting-as-probing}, dialogue \cite{soda}, or evaluation suite \cite{model_behavior}. However, these works typically impose a strong assumption on the teacher LM \cite{gpt-3-labeling-cost}, and require manually constructed set of prompts \cite{i2d2}. Overcoming these limitations, \method generalizes data generation into an off-the-shelf setup, removing the dependence to the teacher model's capability for the target task.

\section{Conclusion}
In this work, we propose \method, a novel framework to distill high-quality paraphrase dataset and model from small, low-quality LMs. We show that by leveraging paraphrastic proximity and critic-guided distillation, \method can empower small LMs to outperform competitive counterparts -- in both performance and controllability, across domains and tasks, without training on human-authored paraphrases. Also, we find that \dataset, the natural byproduct of our method, presents higher fidelity and diversity than order of magnitude larger paraphrase datasets. \method shows a promising direction to rediscover the under-explored capabilities of off-the-shelf language models, by accurately identifying their characteristics and amplifying them.

\section*{Acknowledgements}
This work was funded in part by the DARPA MCS program through NIWC Pacific (N66001-19-2-4031), ONR N00014-24-1-2207, NSF DMS-2134012 and IARPA HIATUS via 2022-22072200003.

\bibliography{anthology,custom}

\appendix
\clearpage
\begin{algorithm}[H]
\caption{Diversity Filter}\label{alg:diversity_filter}
\begin{algorithmic}\small
\Require A set of pairs $\mathcal{P}_\text{in} = \{(x_1, y_1), \cdots, (x_{|P|}, y_{|P|})\}$ generated using the same prefix $c$
\Ensure Filtered set of pairs $\mathcal{P}_\text{out}$\\
$E \gets \emptyset$
\For{$i,j \in \bigl[1, |P|\bigr], i \neq j$} \Comment{search for duplicate pairs}
    \If{$P_{\textit{NLI}}(x_i \Rightarrow x_j) > \tau_\textit{entail}$}
        \State $E \gets E \cup \{(x_i, y_i), (x_j, y_j)\}$
    \ElsIf{$P_{\textit{NLI}}(y_i \Rightarrow y_j) > \tau_\textit{entail}$}
        \State $E \gets E \cup \{(x_i, y_i), (x_j, y_j)\}$
    \EndIf
\EndFor\\
$G \gets (\mathcal{P}_\text{in}, E)$ \Comment{define a graph where nodes are pairs and edges connect duplicate pairs}\\
$S \gets \text{Connected-Components}(G)$\\
$\mathcal{P}_\text{out} \gets \emptyset$
\For{$\mathcal{C} \in S$} \Comment{find the max-entailing pair in each connected component}
    \State $p_\text{out} = \text{argmax}_{(x, y) \in C} P_{\textit{NLI}}(x \Leftrightarrow y)$
    \State $\mathcal{P}_\text{out} \gets \mathcal{P}_\text{out} \cup \{p_\text{out}\}$
\EndFor
\end{algorithmic}
\end{algorithm}

\section{Implementation Details}
\label{sec:implementation_details}
\subsection{Pair Generation}
As noted in Section \ref{sec:dataset_and_model}, we start off using GPT2-XL and BioGPT-large as teacher LM -- all with 1.5B parameters -- generating paraphrases in general and biomedical domain respectively. To sample contextual constraints from these LMs, we use Nucleus Sampling with $\textit{top\_p} = 0.9$ and $\textit{temp} = 1.0$. Additionally, we find that for general domain, generating new-style sentences by prompting GPT2 with a simple prefix (\eg \texttt{New York (CNN) --}) leads to less noisy and more diverse context. For BioGPT, we free-form generate without any prefix given. Throughout the distillation process, we used 4 Quadro RTX 8000 GPUs.

\subsection{Critic Models for Paraphrase Generation}
For semantic equivalence filter, we use Roberta-large-WANLI \cite{wanli}, readily available at HuggingFace transformers \cite{huggingface}. To leave only the highly semantically equivalent pairs of paraphrases, we use $\tau_{\text{semantic}} = 0.75$, discarding all pairs with the bidirectional entailment score below this threshold. For dissimilarity filter, we use $\tau_{\textit{rouge}}= 0.75$ and $\tau_{\textit{TED}} = 12$. We use Stanford CoreNLP library to parse the constituency tree.

Finally, we present the formal algorithm of the diversity filter in Algorithm \ref{alg:diversity_filter}. We first create an undirected graph $G$ where pairs are nodes and edges exist between duplicate pairs, then find the set $S$ of all connected components in $G$. By discarding all but one with the maximal entailment score $p_{\textit{NLI}}(x \Rightarrow y) + p_{\textit{NLI}}(y \Rightarrow x)$ in each component, we remove the duplicate pairs in the candidate pool. As the duplicate pair search with NLI model is parallelizable, the time complexity follows that of the connected component search, \ie $O(|P| + |E|)$ when using DFS-based algorithm \cite{dfs}.

\begin{table}[t]\centering
\resizebox{.48\textwidth}{!}{
    \begin{tabular}{lcccc}\toprule
         \multicolumn{1}{c}{\textbf{Dataset}} & \multicolumn{2}{c}{\textbf{Quora}} & \multicolumn{2}{c}{\textbf{MSCOCO}}\\\cmidrule(lr){2-3}\cmidrule(lr){4-5}
         \multicolumn{1}{c}{\textbf{Model}} & \textit{iBLEU} & \textit{BLEU} & \textit{iBLEU} & \textit{BLEU} \\\midrule
         Lag VAE & 8.73 & 15.52 & 7.69 & 11.63 \\
         CGMH & 9.94 & 15.73 & 7.84 & 11.45 \\
         UPSA & 12.03 & 18.21 & 9.26 & 14.16 \\
         BT & 11.64 & 11.59 & 9.72 & 14.36 \\
         Corruption & 12.32 & 17.97 & 10.32 & 15.60 \\
         ConRPG & 12.68 & 18.31 & 11.17 & 16.98 \\
         MCPG & \underline{13.58} & \underline{24.84} & \underline{11.99} & \underline{20.54} \\
         \model & \textbf{16.40} & \textbf{27.22} & \textbf{13.15} & \textbf{22.75} \\\bottomrule
    \end{tabular}
}
\caption{Experimental results of \model and unsupervised paraphrase generation methods on Quora and MSCOCO. \model consistently outperforms all unsupervised baselines across both benchmarks, in both metrics.}
\label{tab:additional_experiments}
\end{table}


\section{Comparison with Unsupervised Paraphrase Generation Methods}
\label{sec:additional_experiments}
To better understand the effectiveness of \method, we conduct additional experiments that compare \model against state-of-the-art unsupervised methods for paraphrase generation (\ie trained without human-written reference). Following prior works, we use Quora \cite{quora} and MSCOCO \cite{mscoco} datasets repurposed for paraphrase generation. For baselines, we compare against Lag VAE \cite{lag_vae}, CGMH \cite{cgmh}, UPSA \cite{upsa}, BT \cite{paranmt}, Corruption \cite{corruption}, ConRPG \cite{conrpg}, and MCPG \cite{mcpg}. Following past works, we compute and report iBLEU and 4-gram BLEU of each system. 

The results are as shown in Table \ref{tab:additional_experiments}. \model consistently outperforms all unsupervised baselines across both benchmarks, in both metrics.


\begin{figure*}[ht]
    \centering
    \includegraphics[width=.98\textwidth]{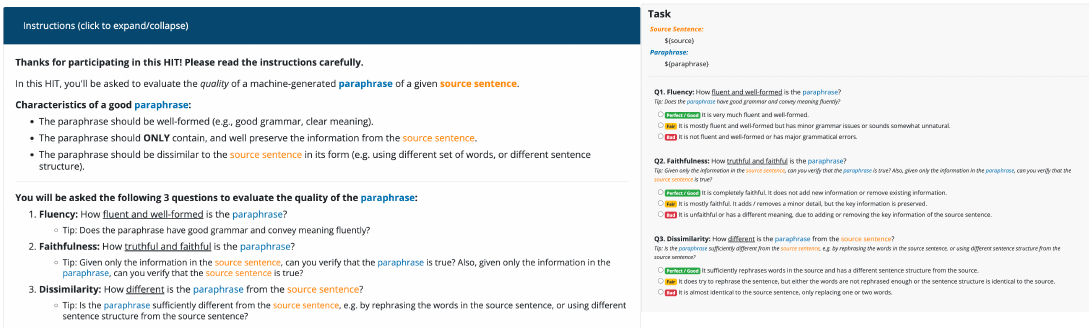}
    \caption{Screenshot of MTurk interface used for the human evaluation of model generated paraphrases.}
    \label{fig:human_eval_template}
\end{figure*}

\section{Generalization to Sentence Summarization}
\label{app:generalization_to_sentence_summarization}
\subsection{Critic Models for Summarization}
In \method, data generation can easily be adapted to sentence summarization, by redefining the filters for the target task (\S\ref{sec:sentence_summarization}). Here, we explain the details of the critic models used for sentence summarization. First, a good summary should be entailed by the original statement without hallucination. Unlike paraphrases, however, summaries allow omitting less important details in the original statement. Therefore, we modify the semantic equivalence filter in paraphrase generation to consider only the unidirectional entailment between $x$ and $y$:
\begin{equation*}
\small
\begin{split}
    f_{\textit{semantic}}(x, y) = \mathbbm{1} \Bigl\{ p_{\textit{NLI}}(x \Rightarrow y) \ge \tau_{\textit{semantic}} \Bigr\}
\end{split}
\end{equation*}

We use the same threshold value as for paraphrase generation, $\tau_{\textit{semantic}} = 0.75$. Next, a desirable summary should be a concise representation of the original statement. We therefore discard all pairs whose compression ratio (\ie the sequence length ratio of $y$ to $x$) is larger than a threshold $\tau_{\textit{comp\_ratio}}$:
\begin{equation*}
\small
\begin{split}
    f_{\textit{comp\_ratio}}(x, y) = \mathbbm{1} \Bigl\{ |y| < |x| \cdot \tau_{\textit{comp\_ratio}} \Bigr\}
\end{split}
\end{equation*}

For our experiments, $\tau_{\textit{comp\_ratio}} = 0.8$. Finally, we employ diversity filter as for paraphrase generation, removing all duplicate \textit{(source, summary)} pairs from the generated dataset:
\begin{equation*}
\small
\begin{split}
    \mathcal{D}_{\textit{T}} = \{(x, y)| & (x, y) \in \mathcal{C}_{\textit{T}}, \\
    & f_{\textit{semantic}} \land f_{\textit{comp\_ratio}} \land f_{\textit{diversity}}(x,y) = 1\}
\end{split}
\end{equation*}

\subsection{DIMSUM and Impossible-T5}
Other than the re-defined filters, we use the same settings as paraphrase generation throughout the distillation pipeline. After self-distillation, we yield a high-quality dataset for sentence summarization (\textbf{D}ataset of \textbf{Im}possible \textbf{Sum}maries, or \textbf{DIMSUM}), with 1.5M sentence-summary pairs across news and biomedical domains. During this process, we also train T5-large into a specialized model for sentence summarization, which we consistently call \textbf{Impossible-T5} as for paraphrase generation.

\section{Human Evaluation Details}
\label{sec:human_evaluation_details}
For human evaluation, we recruit annotators from Amazon Mechanical Turk (MTurk) with an IRB approval, and ensure that all paraphrases are annotated by 6 distinct evaluators with Hit Rate over 99\%. To minimize subjectivity, we use 3-point Likert scale where annotators evaluate the fluency (whether the paraphrase exhibits fleunt language), faithfulness (whether the paraphrase well preserves the content of the original sentence and does not hallucinate), and dissimilarity (whether the paraphrase is sufficiently different from the original statment) of each output. We compensate workers with the hourly wage of \$15. Figure \ref{fig:human_eval_template} shows the actual MTurk interface used for paraphrase evaluation.

\section{Limitations and Future Work}
In this work, we limit our experiments to sentential paraphrasing and summarization tasks. In future works, \method could be applied to a broader range of tasks, \eg translation. To generate a parallel corpus for translation without human supervision, \method could leverage the strong capability of multilingual LMs \cite{xlm, bloom} and cross-lingual filters \cite{xnli}.

\method makes use of a fixed set of filters (\eg off-the-shelf NLI model) to determine which pair qualifies as a high-quality sample. Throughout the distillation pipeline, these filters remain frozen. Although our experiments show that the frozen filters are strong enough to distill a high-quality dataset than state-of-the-art paraphrase corpora, such filters may not always be accessible in wider range of tasks. Hence, future works could improve the framework by learning not only the task model that generates candidate pairs, but also the filter model that scores the plausibility of a given pair. We envision that by co-evolving the task model and filter model throughout the distillation stages, our framework could generalize to more complex problems such as commonsense reasoning, where it is non-trivial to define which pairs qualify as good task example.

As with any distillation technique, \method carries potential risk of amplifying
undesirable properties of language models. While we focus on conditional generation tasks where the output is closely bound to the input, the trained model could inherit the bias and toxicity of its teacher in a more open-ended setting. Nonetheless, \method distills knowledge into a symbolic, textual dataset – which can be interpreted and evaluated, allowing users to intervene in the distillation process and selectively filter which knowledge to be amplified. The inherent transparency of \method, when incorporated with recent techniques for automatic bias detection and reduction, could empower safer knowledge transfer between language models.

\end{document}